\documentclass[letterpaper, 10 pt, conference]{ieeeconf}

\IEEEoverridecommandlockouts

\usepackage{graphics}
\usepackage{graphicx}
\usepackage{cite}
\usepackage{amsmath}
\usepackage{amssymb}
\usepackage{hyperref}
\usepackage{multirow}
\usepackage{xcolor}
\usepackage{tablefootnote}
\usepackage{balance}
\usepackage{orcidlink}

\DeclareMathOperator{\trace}{trace}

\newlength\Myfigwd

\usepackage{multirow}
\usepackage{tabularx}
\usepackage{colortbl}
\usepackage{footnote}
\usepackage{arydshln}

\newcommand{\realfield}[1]{\hbox{I \kern -.25em R}^{#1}}
\newcommand {\mb}[1]{\mathbf{#1}}

\title{\LARGE \bf
Is Image-based Object Pose Estimation Ready to Support Grasping?}

\author{Eric C. Joyce$^{a}$ \orcidlink{0009-0000-4581-4399}, Qianwen Zhao$^{a}$ \orcidlink{0009-0003-0361-1181}, Nathaniel Burgdorfer$^{a}$ \orcidlink{0009-0004-4706-7373}, Long Wang$^{a}$ \orcidlink{0000-0003-3476-6779}, Philippos Mordohai$^{a}$ \orcidlink{0000-0002-9671-4408}
\thanks{$^{a}$Stevens Institute of Technology, Hoboken, NJ 07030, USA. 
        {\tt\small \{ejoyce, qzhao10, nburgdor, lwang4, pmordoha\}@stevens.edu}}%
\thanks{This research was supported in part by NSF Grant CMMI-2138896.}
}

\begin{document}

\maketitle
\thispagestyle{empty}
\pagestyle{empty}

\begin{abstract}
We present a framework for evaluating 6-DoF instance-level object pose estimators, focusing on those that require a single RGB (not RGB-D) image as input. Besides gaining intuition about how accurate these estimators are, we are interested in the degree to which they can serve as the sole perception mechanism for robotic grasping. To assess this, we perform grasping trials in a physics-based simulator, using image-based pose estimates to guide a parallel gripper and an underactuated robotic hand in picking up 3D models of objects. Our experiments on a subset of the BOP (Benchmark for 6D Object Pose Estimation) dataset compare five open-source object pose estimators and provide insights that were missing from the literature.
\end{abstract}

\section{Introduction}\label{sec:intro}
How successfully can object pose estimates made from a single RGB image guide the downstream task of robotic grasping? Most robots (even advanced ones \cite{morgan2022complex,chen2023autobag,qi2023general,zhang2023flowbot++}) rely on RGB-D sensors. Here we investigate the effectiveness of commodity RGB cameras in the instance-level variant of pose estimation, when the 3D shape and appearance of the objects are known a priori. This setting is crucial for robots that operate in industrial and residential environments and should be able to grasp and manipulate known objects given guidance from visual stimuli. Reliance on RGB only (instead of depth sensors) is important for facilitating both indoor and outdoor operation.

Remarkable progress in object pose estimation has been made in the past few years \cite{fan2022deep,hodan2018bop,sahin2020review,sundermeyer2023bop,thalhammer2023challenges}, primarily driven by deep learning and the capability to reduce the so-called \textit{sim2real gap}, enabling end-to-end system training on large amounts of synthetic data with precise ground truth. These systems either predict the pose directly or predict various forms of 2D-3D correspondences which are then fed to a Perspective $n$ Point (PnP) solver to generate the pose (see Section \ref{sec:related}). 

\begin{figure}[!t]
    \centering
    \includegraphics[width=0.9\columnwidth]{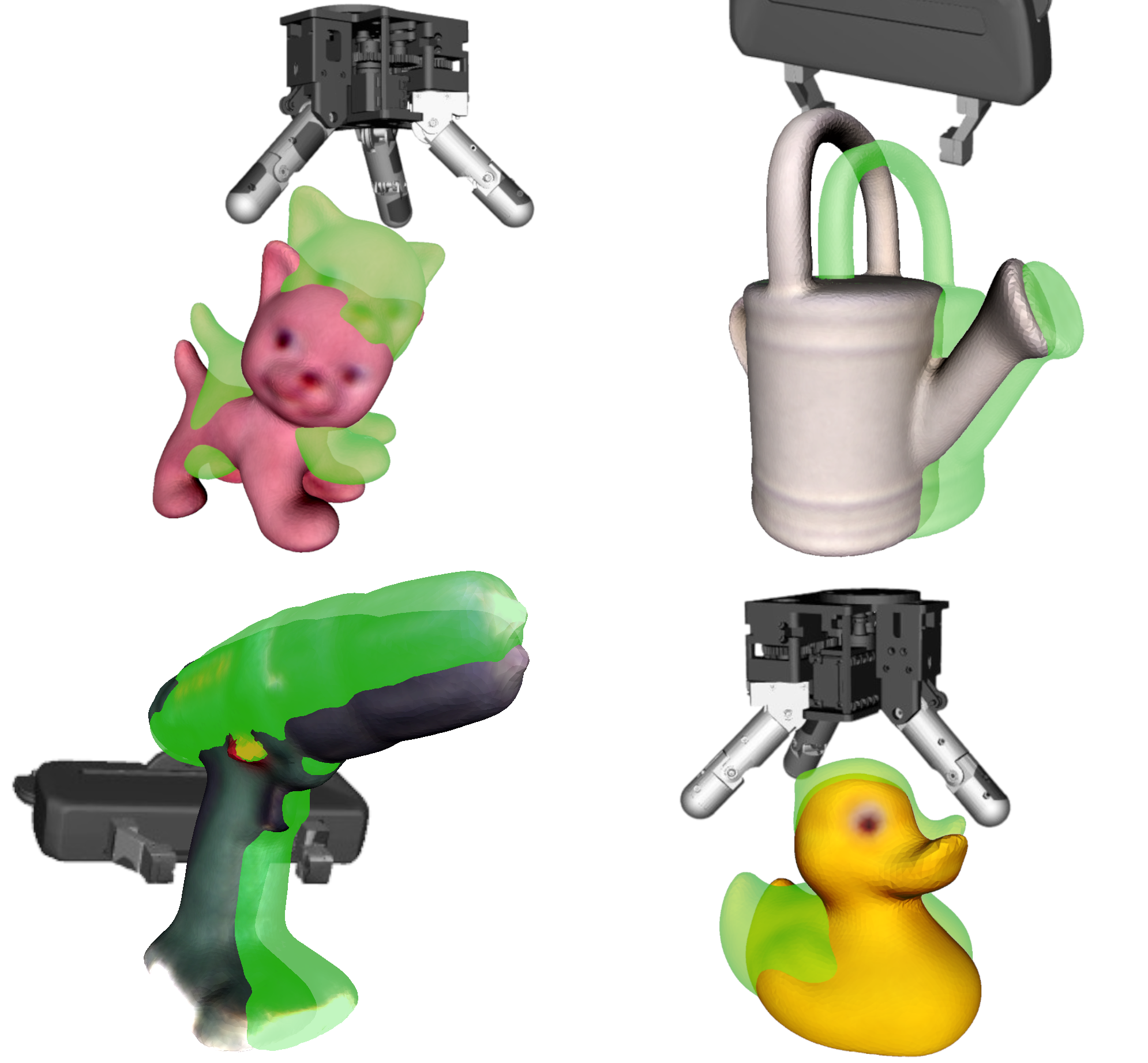}
    \caption{Pose estimates as green overlays and their corresponding ground-truth poses as solid objects. All estimates here measured better than average ADD(-S) and MSSD and yet exhibit significant rotation and translation errors. Our trials attempt to grasp according to the estimated poses, and all estimates seen here were poor enough to cause grasping trial failures.}
    \vspace{-5pt}
    \label{fig:intro_figure_pose_errors_add_mssd}
\end{figure}

Despite comprehensive benchmarks such as the Benchmark for 6D Object Pose Estimation (BOP) \cite{hodan2018bop,sundermeyer2023bop}\footnote{\url{https://bop.felk.cvut.cz/home/}}, the potential for deploying these pose estimators in downstream robotic applications remains unclear. One contributing factor may be that the metrics used to evaluate 6-DoF object pose can sometimes belie subtle geometric errors that cause grasps to fail. Figure~\ref{fig:intro_figure_pose_errors_add_mssd} illustrates some discrepancies obfuscated by ADD(-S) and MSSD, two metrics defined by BOP and found throughout the literature (see Section \ref{sub::metrics}). In assessing how well pose estimates guide different types of robot grippers, we complement the BOP metrics with straightforward rotation and translation errors.

Our framework for evaluating a pose estimator comprises the following steps. For each gripper, we first specify a \textbf{reference grasp} for every object in a dataset. These grasps will be attempted by virtual grippers in open-loop fashion in a simulator. After defining reference grasps for each object-gripper pair, we run the pose estimator on an image containing an object of interest and record the predicted pose, as well as the ground truth that comes with the dataset. We place a virtual model of the object in isolation at the ground-truth pose in the simulator, while the gripper is instructed to grasp it according to the estimated pose, as shown in Fig.~\ref{fig:intro_figure_pose_errors_add_mssd}. We consider a grasping trial successful if the centroid of the object is within a certain tolerance of its target location at the end of the reference grasp (well above the support surface) and remains steadily held for 15 seconds.

This study makes the following assumptions: 3D models of the objects are available; the objects are rigid and of uniform density; the intrinsics of the camera are known. Our experiments are focused on small objects contained in the BOP datasets, and we approximate their weights and friction coefficients with the grippers. The objects are isolated in the simulator. Most of these assumptions can be relaxed as our approach is further developed.

Our experimental results (Section \ref{sec:results}) show that improved predictions across the estimators we study \cite{hodan2020epos,huang2022ncf,tremblay2018deep,su2022zebrapose,wang2021gdr} generally produce higher grasp success rates, though this trend falters for more complex shapes. Certain combinations of estimator, gripper, and object type are more sensitive to error than others.

In summary, the main contributions of our work are:
\begin{itemize}
    \item a framework for evaluating 6-DoF object pose estimation, considering whether pose estimates can be used to guide successful grasping in simulation,
    \item the integration of visual perception on real imagery with a grasping simulator, enabling efficient evaluation of different grippers and grasping success,
    \item an assessment of several representative recent image-based pose estimators that yields new insights on their effectiveness as components of a robotic system.
\end{itemize}

This analysis serves as the foundation for a subsequent study \cite{joyce2025consensus} on learning to predict the success of a robotic grasp before the grasp is attempted.

\section{Related Work}\label{sec:related}
In this section, we review related work on instance-level 6-DoF pose estimation from images and on underactuated robotic hands.

Estimation of 6-DoF poses for known objects from single RGB images has progressed through several phases in the past few years, as shown in surveys \cite{fan2022deep,sahin2020review,thalhammer2023challenges} and the BOP website. The current era of estimation roughly begins with PoseCNN \cite{xiang2017posecnn}, which directly regresses a quaternion and decouples estimates of rotation and translation. 
Robustness to occlusion and handling poses made ambiguous due to symmetry have been the primary concerns motivating developments in deep-learning methods. 
Other research related to 6-DoF pose estimation focuses on improving the performance of PnP \cite{liu2023linear}, on adapting PnP for end-to-end training \cite{chen2022epro}, or on providing statistically defensible bounds for pose estimates \cite{yang2023object}. To emphasize the relevant ideas in this section, we group the paradigms of 6-DoF estimators into sparse, dense, and iterative categories.

\subsection{Sparse Methods}\label{sub::sparse_methods}
Sparse methods predict a handful of key-points from which pose is computed. Both BB8 \cite{rad2017bb8} and DOPE \cite{tremblay2018deep} predict 3D bounding boxes. DOPE learns to predict belief maps about the box's eight corners and vector fields pointing to the predicted object's centroid. 
Inspired by YOLO \cite{redmon2016yolo}, Tekin et al. \cite{tekin2018real} propose a network that performs a single forward pass to predict labels and projections of 3D control points.
Sundermeyer et al. \cite{sundermeyer2018implicit} propose an augmented auto-encoder that learns latent-space representations of objects. These are grouped into a code book used to retrieve rotations, while translation is estimated separately using bounding-box diagonals.

\subsection{Dense Methods}\label{sub::dense_methods}
Though NOCS \cite{wang2019nocs} receives RGB-D as input and predicts poses at the category level, its dense intermediate representation proves useful against occlusion and motivates dense methods for instance-level RGB-only pose estimation.
PVNet \cite{peng2019pvnet} learns to predict a per-pixel vector field for each detected object. Vectors indicate perceived object key-points passed to an uncertainty-weighted version of PnP.
Pix2Pose \cite{park2019pix2pose} learns to predict 3D coordinates for every pixel of a detected object, even when that object is heavily occluded.
EPOS \cite{hodan2020epos} aims at robustness against textureless and symmetric objects by defining objects as sets of fragments. The network learns to predict probabilities for fragments to which a pixel might belong.

Geometry-Guided Direct Regression, or GDR-Net \cite{wang2021gdr}, combines correspondence-based estimation and direct pose regression. GDR-Net generates dense 2D-3D correspondences as intermediate features before directly regressing pose using a learned, patch-based PnP approximator. Wang et al. credit their method's success to thoughtful representations for rotation \cite{zhou2019continuity} and translation \cite{li2019cdpn}, and to a loss function that combines pose and geometry.

ZebraPose \cite{su2022zebrapose} produces dense 2D-3D correspondences by first learning region-specific codes for object vertices. For all objects, vertices are partitioned into iteratively halved regions and assigned a binary feature descriptor to be learned by an encoder-decoder. These codes are ultimately arbitrary, but by training the network in a coarse-to-fine manner, ZebraPose ensures that bits come to represent scales of locality. Once a network has been trained for each object, decoder output for each pixel in a given region of interest is the binary code of the 3D vertex (or neighborhood) corresponding to that 2D pixel. Pose is computed using these correspondences.

Neural Correspondence Field \cite{huang2022ncf} (NCF) samples inside the camera frustum to derive 3D query points rather than pixels for correspondences. This approach aims at mitigating the effects of self-occlusion. NCF then predicts dense 3D-3D correspondences between its query points and points on the object, as well as a signed distance value for each point.

SurfEmb \cite{haugaard2022surfemb} learns to predict dense correspondence distributions over object surfaces without any prior knowledge about object symmetries. This distribution may then be sampled to form and refine pose hypotheses.

\subsection{Iterative Refinement Methods}\label{sub::iter_refine_methods}
DeepIM \cite{li2018deepim} learns to predict a relative pose adjustment that improves upon a given initial pose estimate, and DenseFusion \cite{wang2019densefusion} (an RGB-D method) makes pose refinement a differentiable, iterative process. CosyPose \cite{labbe2020cosypose} and MegaPose \cite{labbe2022megapose} use an iterative, ``render-and-compare" approach to estimate poses. MegaPose simultaneously learns to generalize to object categories. RNNPose \cite{xu2022rnnpose} makes an initial, coarse estimate and improves on it using a recursive refinement module that treats 2D and 3D features separately.

\subsection{Underactuated Robotic Hands and Physics Simulations}
In addition to the widely used Franka Hand \cite{haddadin2022franka}, a parallel gripper, we also simulate grasping trials with a tendon-driven underactuated hand \cite{Aukes2014,Catalano2014,chen2020underactuation,Ciocarlie2009,Ciocarlie2014,Dollar2010,Gosselin2008,Odhner2014,Stuart2017,wang2011highly}. These hands have become appealing for a number of reasons, including their mechanical compliance which allows for a simplified, open-loop control scheme and adapts to object shape variations when grasping.
The low cost and light-weight designs of underactuated hands enable use at scale. Compared to their counterpart, fully-actuated dexterous hands \cite{Jacobsen1986,Loucks1987,Deshpande2013}, underactuated hands can have higher and more realistic tolerance when object pose estimation errors are present.\par 
The underactuated hand in our work is a recent design \cite{chen2020underactuation}, the physics simulation of which has been used previously in deep reinforcement learning \cite{chen2020hardware}. Our grasping trials are performed in MuJoCo \cite{mujoco}.

\section{Method}\label{sec:method}
\subsection{Object Pose Estimation}\label{sub::perception}

Grasping an object requires an estimate of its pose, which can be obtained by any 6-DoF estimator. Given a single RGB image, the estimator predicts a rotation and translation for all instances of known, rigid objects in the scene. Specifically, object $i$ as perceived in image $j$ yields a predicted pose $^{W_{i,j}}\hat{\mb{T}}_{O_{i,j}}$. This notation signifies a rigid transformation from the object's frame $\{O\}$ to the world frame $\{W\}$.

Our interest is in assessing the quality of the estimated pose as applied to the downstream task of grasping. To make this assessment we use the ground-truth poses included in the dataset. For object $i$ in image $j$, the ground-truth pose is $^{W_{i,j}}\mb{T}_{O_{i,j}}$. Poses of symmetric objects may be ambiguous, so measuring error in these cases requires special considerations described in Section \ref{sub::metrics}. Our Physics-based Grasping Simulation module receives the estimated and ground-truth poses from the Object Pose Estimation module.

\subsection{Physics-based Grasping Simulation}\label{sub::simulation}
Physics-based grasping simulation is used to output a binary success score for each pose estimation. For each object-gripper pair, we handcraft a reference grasp plan based on the ground-truth object pose and an open-loop control policy. 

\subsubsection{Parallel and Underactuated Grippers}
The virtual parallel gripper used in our simulator trials is the Franka Hand \cite{haddadin2022franka}, and the underactuated hand used is the design case III presented in \cite{chen2020underactuation}. We anticipate that the more advanced underactuated hand will better tolerate pose error. Comparing their performances will indicate how successfully pose estimators can mitigate this disparity.

\subsubsection{Open-loop Control Policy and Reference Grasps}
\begin{figure}[!t]
    \centering
    \includegraphics[width=0.85\columnwidth]{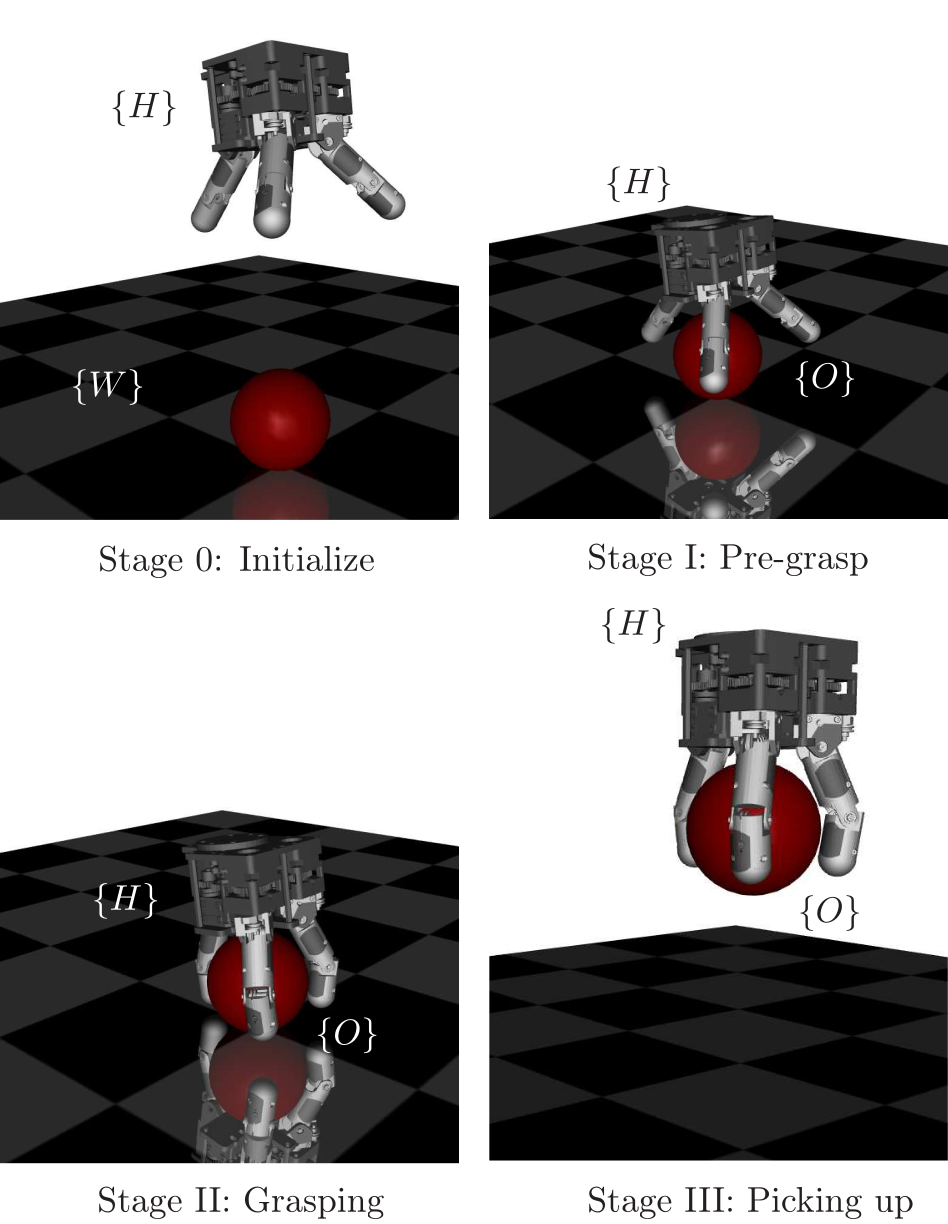}
    \caption{Breakdown of different stages of a simulated grasping task using a simplified open-loop control policy.}
    \vspace{-10pt}
    \label{fig:open_policy}
\end{figure}

We use a simplified, rigid, and open-loop control policy to execute a grasping and picking task for each object. It is termed open-loop because the system does not utilize any sensory feedback, except for an initial object pose estimate to be used in the pre-planned open-loop trajectory. 
Figure~\ref{fig:open_policy} illustrates the four stages of the open-loop control policy. 
The available control actions include the position and orientation of the gripper base (6-DoF) and the single actuator that has one DoF for both grippers. In Stage 0, the gripper is positioned and oriented to an initial pose that is free from collisions with the environment or objects; in Stage I, the gripper is moved to a pre-grasp configuration close to the object; in Stages II and III, the gripper is actuated to close and then to pick up the object.\par

It is worth noting that the most critical part is Stage I, in which the pre-grasp gripper position and orientation are determined by the pose given by an estimator.\par
\begin{figure*}[!h]
\centering
  \includegraphics[width=0.9\textwidth]{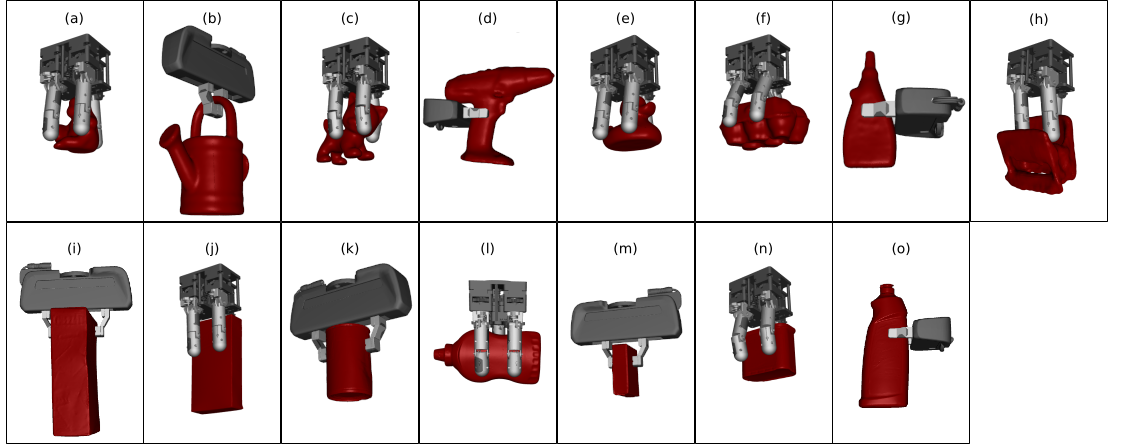}
  \caption{Example reference grasps for selected objects in the LM-O dataset (a-h) and YCB-V dataset (i-o). All grasping trials are attempted with both the parallel gripper and the underactuated hand.}
  \vspace{-10pt}
  \label{fig:reference_grasps}
\end{figure*}

When using the ground-truth object pose to generate the grasp commands for a given object-gripper pair, we term this set of commands a \textbf{reference grasp}. For the experiments shown in Section \ref{sec:results}, we have selected a total of 15 objects from the LM-O and YCB-V datasets.
All selected objects and
one of the two reference grasps for each of them are illustrated in Fig.~\ref{fig:reference_grasps}.

\subsubsection{Generating Grasping Results for Given Object Pose Estimates}
Each pose estimate generated by an estimator may deviate from the ground truth. We capture this deviation and then apply it to a reference grasp as follows.\par
The estimated and the ground-truth pose of the $i^\text{th}$ object in the $j^\text{th}$ image are both described in a shared World, $\{W_{i,j}\}$. 
Therefore, the estimate's deviation is expressed as:
\begin{equation}
    {}^{W_{i,j}}\mb{T}_{\Delta \text{Est.}_j/\text{GT.}} = {}^{W_{i,j}}\hat{\mb{T}}_{O_{i,j}} \; \Big({}^{W_{i,j}}\mb{T}_{O_{i,j,\text{GT}}}\Big)^{-1}
\end{equation}
Then, using the following equations, we rewrite the above pose errors in the physics simulator's world.
\begin{align}
    &    \begin{array}{l}
             {}^{O_{i,j,\text{GT}}}\mb{T}_{\Delta \text{Est.}_j/\text{GT.}} =   \\[6pt]
          \qquad \left({}^{W_{i,j}}\mb{T}_{O_{i,j,\text{GT}}}\right)^{-1} \;\; 
    {}^{W_{i,j}}\mb{T}_{\Delta \text{Est.}_j/\text{GT.}} \;\; {}^{W_{i,j}}\mb{T}_{O_{i,j,\text{GT}}}
    \end{array}
    \label{eqn:pose_change_W1_W2_times2} \\[6pt]
    &    \begin{array}{l}
             {}^{W_\text{Sim.}}\mb{T}_{\Delta \text{Est.}_j/\text{GT.}} =   \\[6pt]
          \qquad \left({}^{O_{i,\text{GT}}}\mb{T}_{W_\text{Sim.}}\right)^{-1} \;\; 
    {}^{O_{i,\text{GT}}}\mb{T}_{\Delta \text{Est.}_j/\text{GT.}} \;\; {}^{O_{i,\text{GT}}}\mb{T}_{W_\text{Sim.}}
    \end{array}
    \label{eqn:pose_change_W1_W2_times3}
\end{align}

Finally, the updated grasp plan can thereby be executed in the physics simulation.
\begin{equation}
   ^{W_\text{Sim.}}\mb{T}_{\text{Plan}_{i, j}} \; = \; {}^{W_\text{Sim.}}\mb{T}_{\Delta \text{Est.}_j/\text{GT.}} \;\; ^{W_\text{Sim.}}\mb{T}_{H_i, \text{ref.}}  
\end{equation}
where a gripper's reference grasp is denoted as $^{W_\text{Sim.}}\mb{T}_{H_i, \text{ref.}}$ and $\{H\}$ is the hand frame. The precise definition of success used in our experiments is provided in Section \ref{sub::metrics}.

\section{Experimental Results}\label{sec:results}
Here, we first introduce the datasets, define the metrics, and review the pose estimators used in this study. Then, we present quantitative results and draw conclusions from them. Please see our video for qualitative results.

\subsection{Datasets}\label{sub::datasets}

The BOP Challenge \cite{sundermeyer2023bop} unifies several datasets for training and evaluating 6-DoF pose estimators. Each dataset includes 3D models of the objects and annotations specifying object symmetries. A dataset contains one or more scenes for training and for testing. Each scene has RGB images, camera intrinsics, and ground-truth 6-DoF poses.

We report experimental results on two of the more popular BOP datasets. YCB-V contains scenes of common household objects and groceries. Its RGB images have (640 $\times$ 480) resolution. Although YCB-V has a total of 21 objects, we limit our experiments to only seven of the least challenging items, following the selection made by the authors of DOPE \cite{tremblay2018deep}. We constrain the other estimators to this same subset for fair comparison. LM-O, also (640 $\times$ 480), is a single scene of eight objects in a cluttered environment. The LM-O objects have more complex shapes than YCB-V, being mostly small toys and handheld tools. We exclude from trials all frames in which target objects have less than 0.5 visibility. Figure~\ref{fig:reference_grasps} shows the shapes of the objects relative to the grippers.

While the dimensions of the objects are precisely captured by the dataset, BOP metadata do not include information on the weights or friction coefficients of objects, which are needed for our simulations. However, their physical details are straightforward to estimate. We also assume that objects are non-deformable and their densities are uniform.

\subsection{Metrics}\label{sub::metrics}

\textbf{Rotation error} $e_{\mb{R}}^{(i, j)}$ and \textbf{translation error} $e_{\mb{t}}^{(i, j)}$ are derived from predicted and true poses, which we define as:
\begin{equation}
    ^{W_{i,j}}\hat{\mb{T}}_{O_{i,j}} = \begin{bmatrix}\mb{\hat{R}} & \mb{\hat{t}} \\
                                                          \mb{0} & 1\end{bmatrix}, \\
    \qquad
    ^{W_{i,j}}\mb{T}_{O_{i,j,\text{GT}}} = \begin{bmatrix}\mb{R} & \mb{t} \\
                                                          \mb{0} & 1\end{bmatrix}
    \label{eqn:T_est_T_gt}
\end{equation}

\noindent To avoid misrepresenting estimates for symmetric objects with ambiguous poses, computation of rotation error considers discrete and continuous symmetries, which are included in BOP metadata. For the former, BOP specifies a set $\mb{S}$ of symmetric rotations, and for the latter, a unit-vector axis of symmetry $\textbf{a}$. Rotation error in the discrete case is determined by the minimizing symmetry:
\begin{equation}
    e_{\mb{R}}^{(i, j)} = \min_{S \in \mb{S}} \arccos\left(\frac{\trace(\mb{\hat{R}}S\mb{R}^\intercal) - 1}{2}\right)
    \label{eqn:e_R_discrete_symmetry}
\end{equation}

\noindent Rotation error in the continuous case is measured as deviation from the axis of symmetry:
\begin{equation}
    e_{\mb{R}}^{(i, j)} = \arccos \big(\textbf{a}^\intercal\mb{\hat{R}}^\intercal\mb{R}^\intercal\mb{\hat{R}}\textbf{a}\big)
    \label{eqn:e_R_continuous_symmetry}
\end{equation}

\noindent \textbf{Maximum Symmetry-Aware Surface Distance} ($e_{\text{MSSD}}$) \cite{hodan2020bop}: This measures prediction misalignment as the single greatest distance between object points in their estimated and in their true poses. $e_{\text{MSSD}}$ is made ``symmetry-aware" by selecting the symmetry that minimizes the greatest distance.

\begin{equation}
    \begin{array}{l}
	e_{\text{MSSD}}\big(\hat{\mathbf{T}}, \mathbf{T}, \mb{S}, \mb{X}\big) = \underset{S \in \mb{S}}{\text{min}} \big( \underset{\textbf{x} \in \mb{X}}{\text{max}}
	\big\Vert \hat{\mathbf{T}}\textbf{x} - \mathbf{T}S\textbf{x}
	\big\Vert_2 \big)
    \end{array}
    \label{eqn:e_MSSD}
\end{equation}

\noindent For an object under consideration, $\mb{S}$ is the set of symmetries, and $\mb{X}$ is the set of vertices.

\noindent \textbf{Maximum Symmetry-Aware Projection Distance} ($e_{\text{MSPD}}$) \cite{hodan2020bop}: This metric behaves similarly to $e_{\text{MSSD}}$ but measures the single greatest distance between pixels of object points projected from predicted and ground-truth poses.

\begin{equation}
    \begin{array}{l}
	e_{\text{MSPD}}\big(\hat{\mathbf{T}}, \mathbf{T}, \mb{S}, \mb{X}\big) = \underset{S \in \mb{S}}{\text{min}} \big( \underset{\textbf{x} \in \mb{X}}{\text{max}}
	\big\Vert \pi( \hat{\mathbf{T}}\textbf{x} ) - \pi(
	\mathbf{T}S\textbf{x} ) \big\Vert_2 \big)
    \end{array}
    \label{eqn:e_MSPD}
\end{equation}

\noindent $\pi(\cdot)$ denotes projection to 2D. The intuition in both $e_{\text{MSSD}}$ and $e_{\text{MSPD}}$ is that we penalize the most egregious misalignment, given the most forgiving symmetry.

\noindent \textbf{Average Distance of Distinguishable Model Points (Symmetric)} ($\text{ADD(-S)}$):
The ADD(-S) metric is still used in the literature, even as the BOP metrics above deprecate it. $\text{ADD(-S)}$ is assigned the Average Distance of Distinguishable Model Points (ADD) or Average Distance of Indistinguishable Model Points (ADI) as applicable, given an object's symmetry. The former averages all distances between corresponding points; the latter seeks each point's nearest neighbor without considering correspondence. The metrics currently advanced by BOP are more rigorous while still making allowances for symmetry.

\noindent \textbf{Grasping Success}: In addition to the above metrics, we introduce a novel measure of grasping success. According to our definition, success requires the object to be within a tolerance of its ideal target location at the end of the reference grasp. Here, we set the tolerance to 5 cm, which means that the distance between the robot hand base and the centroid of the object must be within 5 cm of the target distance 15 seconds after Stage III (in Fig.~\ref{fig:open_policy}). The grasp is specified to end at a sufficient elevation with respect to the table, and any failure to grasp or hold on to the object will be counted as a failure. Unintentional grasps far from the contact points specified by the reference grasp are also likely to be considered failures, depending on the tolerance.

\subsection{Pose Estimators}\label{sub::estimators}

The estimators we have chosen form a representative set of recent works with publicly available code. We use DOPE\footnote{\url{https://github.com/NVlabs/Deep_Object_Pose}}, NCF\footnote{\url{https://github.com/LinHuang17/NCF-code}}, EPOS\footnote{\url{https://github.com/thodan/epos}}, ZebraPose\footnote{\url{https://github.com/suyz526/ZebraPose}}, and GDRNPP\footnote{\url{https://github.com/shanice-l/gdrnpp_bop2022}} as provided, without any further training and without using GDRNPP's refinement module. (GDRNPP is a later iteration of GDR-Net \cite{wang2021gdr}.) In cases where authors offer several sets of weights for the same model, we use the weights that minimize rotation and translation errors on our 15 objects. Although the authors of DOPE provide weights for the YCB-V bottle of bleach, these weights do not yield any successful grasps. We therefore omit this object from DOPE's statistics.

\begin{table*}[!ht]
\centering
\caption{All metrics except the 90th percentile of translation error and success rates are medians.}
\label{tab:ycbv-per-object}
\scalebox{0.9}{
\begin{tabular}{l|ccc|ccc|cc}
\hline
    YCB-V                                                                          &
    \begin{tabular}{@{}c@{}}Rot. Err. \\ (deg)$\downarrow$\end{tabular}            &
    \begin{tabular}{@{}c@{}}Trans. Err. \\ (mm)$\downarrow$\end{tabular}           &
    \begin{tabular}{@{}c@{}}90th perc. \\ Tr. Err. (mm)$\downarrow$\end{tabular}   &
    \begin{tabular}{@{}c@{}}ADD(-S) \\ (mm)$\downarrow$\end{tabular}               &
    \begin{tabular}{@{}c@{}}MSSD \\ (mm)$\downarrow$\end{tabular}                  &
    \begin{tabular}{@{}c@{}}MSPD \\ (pixels)$\downarrow$\end{tabular}              &
    \begin{tabular}{@{}c@{}}Success Rate \\ (Parallel)$\uparrow$\end{tabular}      &
    \begin{tabular}{@{}c@{}}Success Rate \\ (Underactuated)$\uparrow$\end{tabular} \\ \hline\hline
    \textbf{DOPE} \cite{tremblay2018deep} &&&&&&&& \\
    \hspace{10px} Cracker box &
      4.028 &
      17.200 &
      66.040 &
      18.884 &
      24.916 &
      12.927 &
      0.525 &
      0.850 \\ 
    \hspace{10px} Sugar box &
      5.327 &
      27.888 &
      66.712 &
      28.428 &
      32.286 &
      12.607 &
      0.341 &
      0.610 \\ 
    \hspace{10px} Soup can &
      10.213 &
      27.145 &
      62.565 &
      27.355 &
      33.579 &
      11.699 &
      0.096 &
      0.478 \\ 
    \hspace{10px} Mustard bottle &
      26.876 &
      22.736 &
      48.676 &
      27.182 &
      42.364 &
      30.409 &
      0.009 &
      0.549 \\ 
    \hspace{10px} Gelatin box &
      15.987 &
      25.543 &
      47.534 &
      28.078 &
      34.392 &
      20.083 &
      0.569 &
      0.667 \\ 
    \hspace{10px} Potted meat can &
      8.188 &
      16.831 &
      36.586 &
      17.647 &
      24.258 &
      11.610 &
      0.299 &
      0.727 \\ \hline

    \textbf{NCF} \cite{huang2022ncf} &&&&&&&& \\
    \hspace{10px} Cracker box &
      3.313 &
      21.944 &
      42.299 &
      22.654 &
      29.981 &
      11.085 &
      0.364 &
      0.620 \\ 
    \hspace{10px} Sugar box &
      2.755 &
      16.182 &
      28.834 &
      16.328 &
      19.238 &
      9.224 &
      0.795 &
      0.936 \\ 
    \hspace{10px} Soup can &
      12.687 &
      34.566 &
      51.291 &
      35.363 &
      40.744 &
      34.287 &
      0.218 &
      0.406 \\ 
    \hspace{10px} Mustard bottle &
      2.121 &
      23.700 &
      33.123 &
      23.716 &
      26.427 &
      12.215 &
      0.347 &
      0.507 \\ 
    \hspace{10px} Gelatin box &
      6.083 &
      21.466 &
      30.599 &
      21.820 &
      26.947 &
      20.161 &
      0.587 &
      0.787 \\ 
    \hspace{10px} Potted meat can &
      8.157 &
      23.106 &
      49.570 &
      24.127 &
      30.534 &
      23.174 &
      0.099 &
      0.398 \\ 
    \hspace{10px} Bleach cleanser &
      6.013 &
      17.352 &
      44.766 &
      18.044 &
      24.876 &
      11.094 &
      0.117 &
      0.747 \\ \hline

    \textbf{EPOS} \cite{hodan2020epos} &&&&&&&& \\
    \hspace{10px} Cracker box &
      2.038 &
      6.102 &
      20.575 &
      6.440 &
      8.417 &
      6.665 &
      0.743 &
      \textbf{1.000} \\ 
    \hspace{10px} Sugar box &
      1.347 &
      7.784 &
      18.294 &
      8.037 &
      10.436 &
      4.924 &
      0.909 &
      0.992 \\ 
    \hspace{10px} Soup can &
      4.279 &
      8.772 &
      36.759 &
      10.306 &
      13.197 &
      7.152 &
      0.347 &
      0.742 \\ 
    \hspace{10px} Mustard bottle &
      3.205 &
      4.236 &
      7.544 &
      5.601 &
      9.076 &
      7.255 &
      0.407 &
      0.993 \\ 
    \hspace{10px} Gelatin box &
      1.487 &
      7.824 &
      26.228 &
      7.828 &
      8.897 &
      3.427 &
      0.920 &
      0.933 \\ 
    \hspace{10px} Potted meat can &
      1.978 &
      9.159 &
      40.335 &
      9.319 &
      11.346 &
      4.747 &
      0.663 &
      0.890 \\ 
    \hspace{10px} Bleach cleanser &
      3.782 &
      9.955 &
      57.802 &
      11.355 &
      17.824 &
      9.921 &
      0.090 &
      0.723 \\ \hline

    \textbf{GDRNPP} \cite{liu2022gdrnpp_bop} &&&&&&&& \\
    \hspace{10px} Cracker box &
      2.442 &
      9.364 &
      14.388 &
      9.872 &
      13.766 &
      7.257 &
      0.850 &
      \textbf{1.000} \\ 
    \hspace{10px} Sugar box &
      1.396 &
      5.492 &
      10.136 &
      5.629 &
      7.338 &
      4.342 &
      \textbf{1.000} &
      \textbf{1.000} \\ 
    \hspace{10px} Soup can &
      2.727 &
      6.986 &
      15.101 &
      7.065 &
      8.701 &
      6.240 &
      0.535 &
      0.918 \\ 
    \hspace{10px} Mustard bottle &
      2.913 &
      4.607 &
      7.651 &
      4.828 &
      7.846 &
      5.275 &
      0.907 &
      \textbf{1.000} \\ 
    \hspace{10px} Gelatin box &
      7.181 &
      5.076 &
      27.415 &
      6.137 &
      10.944 &
      7.027 &
      0.813 &
      \textbf{1.000} \\ 
    \hspace{10px} Potted meat can &
      1.663 &
      4.126 &
      21.308 &
      4.254 &
      5.551 &
      3.697 &
      0.878 &
      0.928 \\ 
    \hspace{10px} Bleach cleanser &
      2.985 &
      8.606 &
      36.355 &
      9.009 &
      12.740 &
      8.454 &
      0.273 &
      0.773 \\ \hline

    \textbf{ZebraPose} \cite{su2022zebrapose} &&&&&&&& \\
    \hspace{10px} Cracker box &
      1.551 &
      8.791 &
      13.118 &
      8.841 &
      12.126 &
      6.199 &
      0.856 &
      \textbf{1.000} \\ 
    \hspace{10px} Sugar box &
      1.573 &
      6.899 &
      14.527 &
      6.997 &
      9.028 &
      5.204 &
      0.925 &
      \textbf{1.000} \\ 
    \hspace{10px} Soup can &
      1.967 &
      5.123 &
      16.242 &
      5.405 &
      6.855 &
      5.804 &
      0.620 &
      0.918 \\ 
    \hspace{10px} Mustard bottle &
      2.863 &
      3.719 &
      11.034 &
      4.721 &
      8.415 &
      7.026 &
      0.547 &
      0.980 \\ 
    \hspace{10px} Gelatin box &
      1.392 &
      5.296 &
      17.733 &
      5.363 &
      6.825 &
      3.261 &
      0.960 &
      \textbf{1.000} \\ 
    \hspace{10px} Potted meat can &
      1.580 &
      8.278 &
      21.112 &
      8.285 &
      9.313 &
      4.990 &
      0.796 &
      0.961 \\ 
    \hspace{10px} Bleach cleanser &
      2.862 &
      8.250 &
      55.863 &
      8.758 &
      12.473 &
      7.957 &
      0.203 &
      0.763 \\ \hline
\end{tabular}
}
\end{table*}

\begin{table*}[!ht]
\centering
\caption{All metrics except the 90th percentile of translation error and success rates are medians. Note that it is not possible to grasp the egg box object using the parallel gripper, regardless of the quality of the pose estimate.}
\label{tab:lmo-per-object}
\scalebox{0.9}{
\begin{tabular}{l|ccc|ccc|cc}
\hline
    LM-O                                                                           &
    \begin{tabular}{@{}c@{}}Rot. Err. \\ (deg)$\downarrow$\end{tabular}            &
    \begin{tabular}{@{}c@{}}Trans. Err. \\ (mm)$\downarrow$\end{tabular}           &
    \begin{tabular}{@{}c@{}}90th perc. \\ Tr. Err. (mm)$\downarrow$\end{tabular}   &
    \begin{tabular}{@{}c@{}}ADD(-S) \\ (mm)$\downarrow$\end{tabular}               &
    \begin{tabular}{@{}c@{}}MSSD \\ (mm)$\downarrow$\end{tabular}                  &
    \begin{tabular}{@{}c@{}}MSPD \\ (pixels)$\downarrow$\end{tabular}              &
    \begin{tabular}{@{}c@{}}Success Rate \\ (Parallel)$\uparrow$\end{tabular}      &
    \begin{tabular}{@{}c@{}}Success Rate \\ (Underactuated)$\uparrow$\end{tabular} \\ \hline\hline
    \textbf{EPOS} \cite{hodan2020epos} &&&&&&&& \\
    \hspace{10px} Ape &
      6.276 &
      24.997 &
      66.190 &
      24.702 &
      28.742 &
      6.356 &
      0.043 &
      0.389 \\ 
    \hspace{10px} Can &
      4.553 &
      20.478 &
      62.804 &
      20.981 &
      27.774 &
      6.547 &
      0.714 &
      0.794 \\ 
    \hspace{10px} Cat &
      13.349 &
      30.751 &
      76.806 &
      31.622 &
      42.467 &
      8.635 &
      0.073 &
      0.452 \\ 
    \hspace{10px} Drill &
      3.479 &
      15.970 &
      56.227 &
      16.457 &
      21.863 &
      6.482 &
      0.528 &
      0.494 \\ 
    \hspace{10px} Duck &
      9.448 &
      12.004 &
      37.394 &
      13.912 &
      19.477 &
      7.341 &
      0.314 &
      0.571 \\ 
    \hspace{10px} Egg box &
      39.309 &
      82.298 &
      918.043 &
      38.115 &
      205.819 &
      87.161 &
      - &
      0.180 \\ 
    \hspace{10px} Glue &
      6.508 &
      29.666 &
      88.713 &
      12.461 &
      36.163 &
      7.564 &
      0.405 &
      0.587 \\ 
    \hspace{10px} Hole-puncher &
      5.316 &
      22.048 &
      45.440 &
      22.333 &
      27.814 &
      6.759 &
      0.267 &
      0.314 \\ \hline

    \textbf{GDRNPP} \cite{liu2022gdrnpp_bop} &&&&&&&& \\
    \hspace{10px} Ape &
      3.948 &
      9.604 &
      22.075 &
      9.819 &
      12.292 &
      4.539 &
      0.142 &
      0.821 \\ 
    \hspace{10px} Can &
      3.405 &
      11.347 &
      23.162 &
      11.931 &
      16.127 &
      5.250 &
      0.888 &
      0.949 \\ 
    \hspace{10px} Cat &
      3.948 &
      13.426 &
      30.240 &
      13.675 &
      17.509 &
      4.134 &
      0.274 &
      0.847 \\ 
    \hspace{10px} Drill &
      3.112 &
      11.156 &
      26.569 &
      11.733 &
      16.898 &
      5.254 &
      0.590 &
      0.708 \\ 
    \hspace{10px} Duck &
      7.659 &
      19.038 &
      31.551 &
      19.900 &
      24.307 &
      6.008 &
      0.308 &
      0.385 \\ 
    \hspace{10px} Egg box &
      6.686 &
      51.891 &
      694.946 &
      20.611 &
      198.821 &
      84.249 &
      - &
      0.286 \\ 
    \hspace{10px} Glue &
      6.214 &
      14.888 &
      45.016 &
      6.980 &
      19.166 &
      6.670 &
      0.787 &
      0.951 \\ 
    \hspace{10px} Hole-puncher &
      4.431 &
      22.042 &
      39.907 &
      22.082 &
      26.724 &
      6.057 &
      0.171 &
      0.181 \\ \hline

    \textbf{ZebraPose} \cite{su2022zebrapose} &&&&&&&& \\
    \hspace{10px} Ape &
      3.613 &
      8.398 &
      18.365 &
      8.480 &
      10.533 &
      4.981 &
      0.219 &
      0.825 \\ 
    \hspace{10px} Can &
      2.958 &
      5.752 &
      12.421 &
      6.783 &
      9.723 &
      4.349 &
      0.944 &
      0.983 \\ 
    \hspace{10px} Cat &
      3.520 &
      10.338 &
      24.029 &
      10.978 &
      14.901 &
      3.940 &
      0.390 &
      0.878 \\ 
    \hspace{10px} Drill &
      2.522 &
      9.326 &
      20.680 &
      9.546 &
      13.089 &
      4.936 &
      0.669 &
      0.792 \\ 
    \hspace{10px} Duck &
      7.423 &
      7.293 &
      14.117 &
      8.742 &
      13.705 &
      5.902 &
      0.596 &
      0.519 \\ 
    \hspace{10px} Egg box &
      5.199 &
      23.191 &
      1041.358 &
      10.242 &
      177.739 &
      81.047 &
      - &
      0.632 \\ 
    \hspace{10px} Glue &
      4.112 &
      11.980 &
      32.020 &
      5.018 &
      14.751 &
      4.769 &
      0.882 &
      0.958 \\ 
    \hspace{10px} Hole-puncher &
      4.271 &
      9.562 &
      21.111 &
      10.003 &
      13.803 &
      6.150 &
      0.601 &
      0.684 \\ \hline
\end{tabular}
}
\vspace{-10pt}
\end{table*}

\begin{table*}[!ht]
\centering
\caption{Select examples of areas under the curve when grasp failure rate is plotted as a function of each increasing error. Perfect performance for a given estimator, object, and gripper leaves zero area under the curve.}
\label{tab:auc-per-object-selects}
\scalebox{0.9}{
\begin{tabular}{l|cccc|cccc}
\hline
          & \multicolumn{4}{c}{\textbf{Parallel}} & \multicolumn{4}{c}{\textbf{Underactuated}}\\
    YCB-V                                                                       &
    \begin{tabular}{@{}c@{}}AUC \\ Rot.Err.$\downarrow$\end{tabular}            &
    \begin{tabular}{@{}c@{}}AUC \\ Trans. Err.$\downarrow$\end{tabular}         &
    \begin{tabular}{@{}c@{}}AUC \\ ADD(-S)$\downarrow$\end{tabular}             &
    \begin{tabular}{@{}c@{}}AUC \\ MSSD $\downarrow$\end{tabular}               &
    \begin{tabular}{@{}c@{}}AUC \\ Rot.Err.$\downarrow$\end{tabular}            &
    \begin{tabular}{@{}c@{}}AUC \\ Trans. Err.$\downarrow$\end{tabular}         &
    \begin{tabular}{@{}c@{}}AUC \\ ADD(-S)$\downarrow$\end{tabular}             &
    \begin{tabular}{@{}c@{}}AUC \\ MSSD $\downarrow$\end{tabular}               \\ \hline\hline

    \textbf{NCF} \cite{huang2022ncf} &&&&&&&& \\
    \hspace{10px} Sugar box &
      4.575 &
      2.823 &
      2.750 &
      \textbf{2.568} &
      0.834 &
      0.242 &
      \textbf{0.239} &
      0.285 \\
    \hspace{10px} Soup can &
      80.041 &
      69.513 &
      70.126 &
      \textbf{65.826} &
      58.321 &
      33.700 &
      34.097 &
      \textbf{30.760} \\
    \hspace{10px} Mustard bottle &
      45.174 &
      41.231 &
      41.235 &
      \textbf{38.018} &
      38.256 &
      20.275 &
      \textbf{19.680} &
      20.770 \\ \hline

    \textbf{EPOS} \cite{hodan2020epos} &&&&&&&& \\
    \hspace{10px} Sugar box &
      1.829 &
      1.307 &
      1.238 &
      \textbf{0.787} &
      0.009 &
      0.029 &
      0.018 &
      \textbf{0.004} \\
    \hspace{10px} Soup can &
      48.646 &
      31.658 &
      \textbf{29.760} &
      29.811 &
      6.201 &
      9.421 &
      \textbf{3.689} &
      3.702 \\
    \hspace{10px} Mustard bottle &
      38.913 &
      43.484 &
      31.722 &
      \textbf{29.632} &
      0.128 &
      \textbf{0.003} &
      \textbf{0.003} &
      \textbf{0.003} \\ \hline

    \textbf{GDRNPP} \cite{liu2022gdrnpp_bop} &&&&&&&& \\
    \hspace{10px} Sugar box &
      \textbf{0} &
      \textbf{0} &
      \textbf{0} &
      \textbf{0} &
      \textbf{0} &
      \textbf{0} &
      \textbf{0} &
      \textbf{0} \\
    \hspace{10px} Soup can &
      29.714 &
      13.573 &
      \textbf{13.448} &
      13.587 &
      1.327 &
      0.516 &
      0.513 &
      \textbf{0.475} \\
    \hspace{10px} Mustard bottle &
      \textbf{4.029} &
      13.178 &
      10.422 &
      10.064 &
      \textbf{0} &
      \textbf{0} &
      \textbf{0} &
      \textbf{0} \\ \hline

    \textbf{ZebraPose} \cite{su2022zebrapose} &&&&&&&& \\
    \hspace{10px} Sugar box &
      1.247 &
      0.302 &
      0.302 &
      \textbf{0.296} &
      \textbf{0} &
      \textbf{0} &
      \textbf{0} &
      \textbf{0} \\
    \hspace{10px} Soup can &
      31.318 &
      8.788 &
      \textbf{8.769} &
      9.292 &
      5.525 &
      0.560 &
      0.547 &
      \textbf{0.453} \\
    \hspace{10px} Mustard bottle &
      46.768 &
      32.963 &
      28.053 &
      \textbf{26.131} &
      2.610 &
      \textbf{0.017} &
      \textbf{0.017} &
      \textbf{0.017} \\ \hline

    LM-O                                                                          &
    &&&&&&&\\ \hline\hline
    \textbf{EPOS} \cite{hodan2020epos} &&&&&&&& \\
    \hspace{10px} Can &
      19.251 &
      8.843 &
      \textbf{8.184} &
      8.307 &
      13.391 &
      5.318 &
      \textbf{4.468} &
      4.607 \\
    \hspace{10px} Drill &
      41.827 &
      21.772 &
      \textbf{21.241} &
      21.430 &
      40.100 &
      17.977 &
      17.919 &
      \textbf{17.555} \\
    \hspace{10px} Duck &
      60.396 &
      45.463 &
      45.328 &
      \textbf{45.180} &
      39.468 &
      \textbf{34.914} &
      36.749 &
      37.220 \\ \hline

    \textbf{GDRNPP} \cite{liu2022gdrnpp_bop} &&&&&&&& \\
    \hspace{10px} Can &
      7.158 &
      2.645 &
      \textbf{2.458} &
      2.792 &
      1.453 &
      0.252 &
      0.275 &
      \textbf{0.202} \\
    \hspace{10px} Drill &
      34.231 &
      22.092 &
      20.166 &
      \textbf{19.229} &
      26.539 &
      8.458 &
      \textbf{8.167} &
      8.763 \\
    \hspace{10px} Duck &
      68.248 &
      55.686 &
      \textbf{55.438} &
      56.739 &
      47.296 &
      51.029 &
      49.592 &
      \textbf{47.107} \\ \hline

    \textbf{ZebraPose} \cite{su2022zebrapose} &&&&&&&& \\
    \hspace{10px} Can &
      3.692 &
      \textbf{0.386} &
      0.441 &
      0.719 &
      0.312 &
      \textbf{0.014} &
      \textbf{0.014} &
      \textbf{0.014} \\
    \hspace{10px} Drill &
      23.268 &
      14.383 &
      14.722 &
      \textbf{12.824} &
      12.965 &
      4.278 &
      4.355 &
      \textbf{4.182} \\
    \hspace{10px} Duck &
      36.898 &
      \textbf{23.623} &
      24.266 &
      25.465 &
      \textbf{44.644} &
      45.165 &
      44.899 &
      44.917 \\ \hline
\end{tabular}
}
\vspace{-10pt}
\end{table*}

\subsection{Quantitative Results}\label{sub:quant}

Table \ref{tab:ycbv-per-object} reports per-object median errors and average grasp success rates for YCB-V. We take the expected behavior observed here as validation of our study. Both grippers perform better on YCB-V than on LM-O (compare Table \ref{tab:lmo-per-object}), and we attribute this to the relatively simple shapes of the YCB-V objects: prismatic (three boxes), cylindrical (soup can), and ergonomic (two squeeze bottles). Though the parallel gripper lags behind the underactuated gripper, grasping success for both tends to increase as errors decrease, and the least challenging objects saturate first, namely the prisms. On these objects, both grippers can tolerate some rotation and translation error. Objects for which we measure competitive \textit{median} errors may nevertheless remain challenging for the parallel gripper if an estimator's high-end (90th percentile) translation errors are large. When we observe low geometric error and low success rates, as for the parallel gripper on non-prisms, we may conclude that the gripper has become the limiting factor.

The general alignment between reduced error and increased grasp success becomes less reliable in the more challenging LM-O set, summarized in Table \ref{tab:lmo-per-object}. LM-O contains no prisms or cylinders. The parallel gripper's performance on ``free-form" objects such as Ape, Cat, and Duck indicates that it is not suited for these objects. Even as pose estimates improve, the concavity and curvature of these small figurines make parallel grasps highly sensitive to error.

\begin{figure}[!b]
    \centering
    \includegraphics[width=0.9\columnwidth]{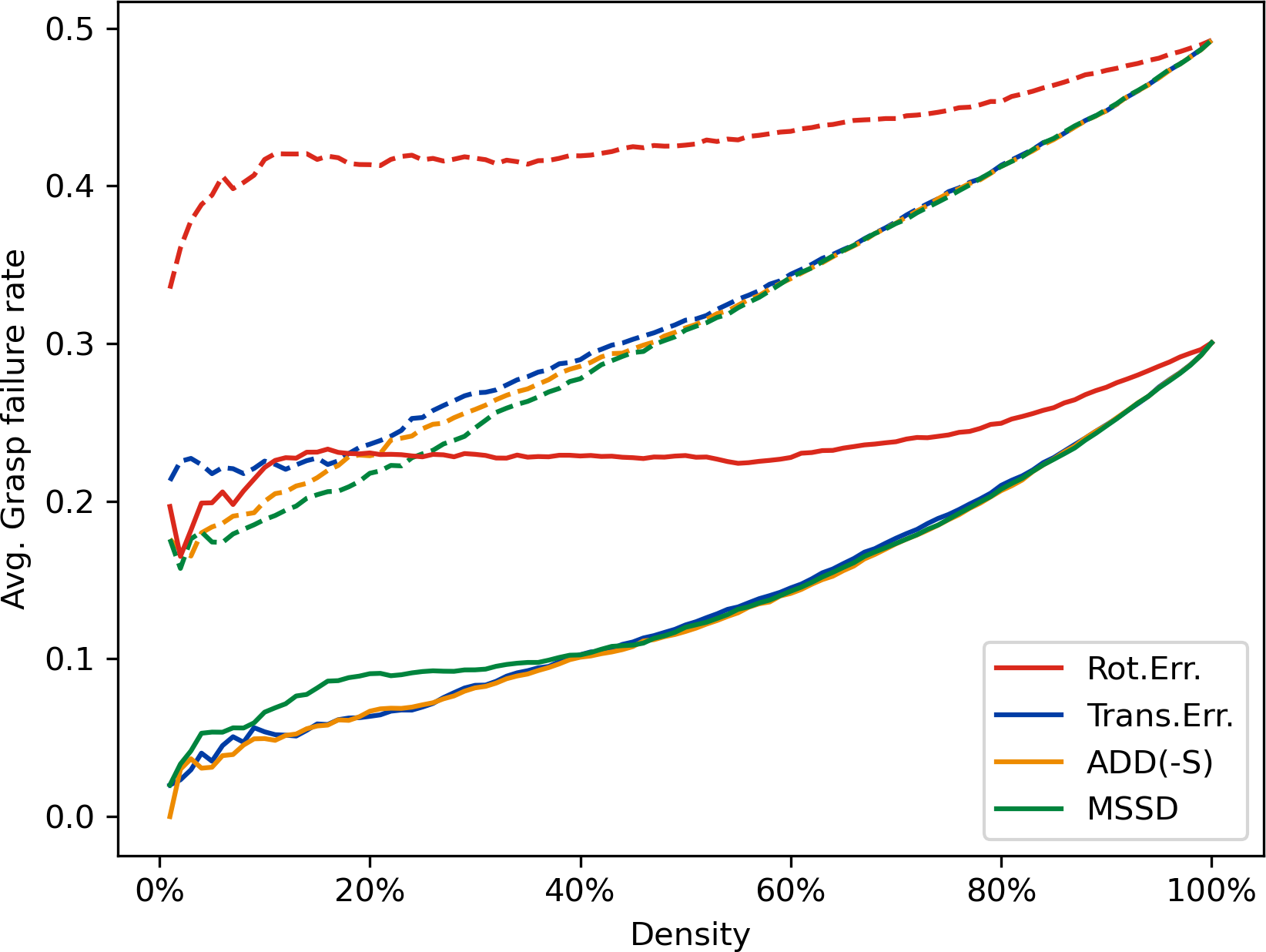}
    \vspace{-5pt}
    \caption{Cumulative distribution curves for grasp \textit{failure} rate as a function of our four metrics. These curves average together all objects, for all estimators. Dashed lines are for the parallel gripper, while solid lines are for the underactuated hand. The metric with least area under its curve is the strongest predictor for grasp success. Here we see the overall superiority of the underactuated hand, the pronounced tolerance to rotation error, and the correlation between translation error and the two BOP metrics.}
    \label{fig:gripper-CDF}
\end{figure}

Figure~\ref{fig:gripper-CDF} plots cumulative grasp \textit{failure} rates as a function of each of our four \textit{increasing} metrics. Each area under the curve (AUC) indicates the predictive power of that metric for that gripper. An ideal predictor's cumulative distribution should include all the \textit{successes} first as we admit more trial results and therefore correspond to the lowest possible AUC. A meaningless predictor is essentially random and would approach a horizontal line at the average failure rate for all trials. In general, rotation error is the least informative predictor of failure.

Table \ref{tab:auc-per-object-selects} reports select AUCs for illustrative estimator-object pairs. As grasp success on prisms saturates, their AUCs drop to zero: when performance is perfect, there is no failure to indicate.
Analysis of AUCs reveals that grasp failure for the majority of our objects is determined by translation error. Since ADD(-S) and MSSD are strongly correlated with translation error, their predictive powers are similar.
Decomposing the translation errors across all estimators and objects, we can see that at least 80\% of $e_{\mb{t}}^{(i, j)}$ occurs along the viewing direction, orthogonal to the camera's image plane. This is to be expected, given the lack of an input depth channel. Rotation seems especially insignificant for cylinders, which makes sense given that rotations around their axis of symmetry does not affect grasp. Ergonomic objects and the parallel gripper exhibit sensitivity to rotation. These grasps fail when closure of the parallel pincers does not align with the objects' minor axes. Recall that in the physics simulator, objects are non-deformable. In real life, ergonomic objects could be squeezed, and misaligned parallel grasps might succeed. The underactuated hand is sensitive to rotation on free-form objects. 
The rotation errors and the arbitrariness in objects' 3D shapes lead to circumstances in which underactuated fingers slide away from stable force closure configurations, causing object ejection \cite{birglen2003force}. 

\section{Conclusions}\label{sec:conclusion}
In this paper, for the first time we have attempted to measure how successful a robot hand would be in grasping objects following an open-loop policy based on pose estimates from an RGB image. Whether image-based object pose estimation is ready to support grasping depends on which gripper is used and on the shape of the target object. Our experiments with several object-pose estimators demonstrate that errors are shrinking as the estimators improve, but that a gripper unsuited to its target will become an impediment regardless of the quality of the estimate. We have seen that even poor pose estimates may be tolerated for prismatic objects, but that intricate shapes demand greater accuracy and dexterity. We conclude that a state of the art, competitive pose estimator is necessary, and that the simpler, parallel gripper may serve if the only objects to be grasped are prisms. The underactuated hand has higher tolerance for rotation errors, due to its larger working area, and can succeed where the parallel gripper fails.

{\small
\bibliographystyle{IEEEtran}
\bibliography{bib/bibliography,%
bib/hand_designs}
}
\balance
\end{document}